\documentclass[preprint, 3p, twocolumn]{elsarticle}

\usepackage{epsfig}
\usepackage{amssymb}
\makeatletter
\newif\if@restonecol
\makeatother

\usepackage{lineno}
\usepackage[linesnumbered,ruled]{algorithm2e}
\usepackage{float}
\usepackage{amsmath}

\AtBeginDocument{%
  \providecommand\BibTeX{{%
    \normalfont B\kern-0.5em{\scshape i\kern-0.25em b}\kern-0.8em\TeX}}}

\begin{document}

\begin{frontmatter}



\title{The reinforcement learning-based approach for the adaptive speed regulation on a metallurgical pickling line}

\author{Anna Bogomolova\corref{cor1}}
\ead{aiu.bogomolova@severstal.com}

\author{Kseniia Kingsep}
\ead{ka.kingsep@severstal.com}

\author{Boris Voskresenskii}
\ead{bs.voskresenskii@severstal.com}

\cortext[cor1]{{Corresponding author}}
\affiliation{organization={Severstal Digital},
            addressline={Klary Tsetkin Street, 2},
            city={Moscow},
            postcode={127299},
            country={Russia}}

% \author[inst1]{Author One}

% \affiliation[inst1]{organization={Department One},%Department and Organization
%             addressline={Address One}, 
%             city={City One},
%             postcode={00000}, 
%             state={State One},
%             country={Country One}}

% \author[inst2]{Author Two}
% \author[inst1,inst2]{Author Three}

% \affiliation[inst2]{organization={Department Two},%Department and Organization
%             addressline={Address Two}, 
%             city={City Two},
%             postcode={22222}, 
%             state={State Two},
%             country={Country Two}}

\begin{abstract}
We present a holistic data-driven approach to the problem of productivity increase on the example of a metallurgical pickling line. The proposed approach combines mathematical modeling as a base algorithm and a Reinforcement Learning system implemented such as to enhance the performance of the base algorithm by multiple criteria while also meeting safety and reliability requirements and taking into account the unexpected volatility of certain technological processes. We demonstrate how Deep Q-Learning can be applied to a real-life task in heavy industry, resulting in significant improvement of previously existing automation systems. The problem of input data scarcity is solved by a two-step combination of LSTM and CGAN, which helps to embrace both the tabular representation of the data and its sequential properties. Offline RL training, a necessity in this setting, has become possible through the sophisticated probabilistic kinematic environment.
\end{abstract}


% \begin{highlights}
% \item The control problem for the metallurgical pickling line is solved using the Reinforcement learning framework.
% \item A combination of mathematical modeling with the reinforcement learning approach was applied to meet the technological requirements and take advantage of the data-driven approach.
% \item Proved economic efficiency of the suggested approach is shown.
% \item Synthetic data based on the combination of CGAN and LSTM was generated to produce a sufficient amount of input data, such as a sequence of steel strips parameters.
% \end{highlights}

\begin{keyword}
smart manufacture \sep deep-reinforcement learning \sep CGAN \sep control systems
% %% PACS codes here, in the form: \PACS code \sep code
% %% MSC codes here, in the form: \MSC code \sep code
% %% or \MSC[2008] code \sep code (2000 is the default)
\end{keyword}

\end{frontmatter}

% \linenumbers

\section{Introduction}
A pickling line is a continuously-acting line where steel strips processed in a hot-rolling mill are exposed to the removal of iron oxides from their surface. The pickling process is required between hot rolling and any further treatments (e.g. hot deep galvanizing, cold rolling, or annealing) and therefore may become a bottleneck in the production process. Meanwhile, the direct increase in the production speed can negatively affect the quality of the final product, if the mill scale is not properly removed. During the pickling process, steel strips are transported through the baths filled with a concentrated acid solution to remove the mill scale defects formed during the hot-rolling process. Unsuitable conditions of the pickling process (the concentration of the acid solution, the baths temperature, and the line speed \cite{hudson1982effect}) result in incomplete removal of scale from the steel surface, namely under-pickling, and require the re-processing. On the other hand, over-pickling can lead to erosion and has a more sensible effect on steel quality. Therefore, there is a trade-off between high productivity and quality of the steel strips. Most of the research papers done so far try to solve the trade-off in favor of satisfying quality requirements, by finding optimal speed line values for the different combinations of acid concentrations and temperature ranges \cite{colla2012prediction, machon2010flsom} or by controlling the acid concentration in range assured the proper conditions for the pickling reaction \cite{kittisupakorn2009neural, noh2015acid, sohlberg2007control, daosud2005neural}. However, to our knowledge, no one of these publications takes into account the sophisticated design of the pickling line.

\begin{figure*}[tb]
  \centering
  \includegraphics[width=\linewidth]{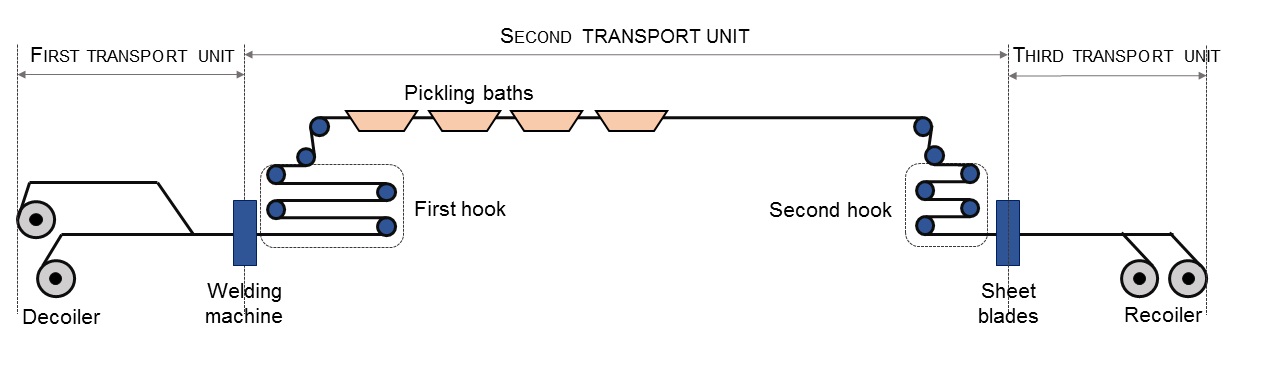}
  \caption{Scheme of picking line}
  \label{fig:schema}
\end{figure*}

While the topology of lines can vary with steel mill, the technological procedure responsible for the continuous operation \cite{simeonov2010acid} of the pickling lines is very similar and affects significantly the control strategy of the line. To make the process continuous, the steel strips are first welded lengthwise at the outlet of the first transport unit (FTU) then fed through a pickling bath (the second transport unit, STU), and finally cut down at the inlet of the third unit (TTU) \cite{tytus1925metal} (Figure \ref{fig:schema}). While the FTU and TTU can either move or stop (to make the welding or cutting process to be possible), the STU has to be always in motion, because the line stoppages lead to steel defects. Therefore, the control strategy for the STU should also cover the logic of continuous operation in conditions when FTU stops for welding and provides less input or TTU stops for cutting and can not let the steel strips out.

In the current publication, we deal with the premise that pickling defects are rare and demand in productivity is high. Given this assumption, we propose an approach in which the constraints formed by quality requirements are satisfied by designing the speed constraints, a table where for some strip width/thickness/steel grade combinations the relevant STU speed range is defined, and the entire problem is solved as a pure optimal control problem with some associated constraints. There are two main components of control that can be applied to the current task, the control of strip sequence at the inlet of the pickling line formulated as a scheduling problem \cite{priore2001review} and the control of speed line at every time moment. While the scheduling approach is more frequently used in manufacturing systems, in our case we follow the production requirements and presume that the sequence of steel strips entering the pickling line is predetermined and unchangeable. Finally, when solving the speed line control problem, a decision is made on the speed of the STU, which corresponds to the boundaries from the speed table, takes into account the FTU and the TTU states, and leads to maximum performance over a long time horizon. In such formulation, the task seems to be perfectly fit the range of problems successfully solved with the Reinforcement Learning (RL) framework.

The main advantage of RL algorithms comes from the ability to construct many potential scenarios considering different actions applied to the system, to evaluate each of them \cite{sutton2018reinforcement}, and, finally, to choose the one that results in desired  maximization/minimization of the objective function. Moreover, they also allow assessing a long-term effect of consequently taken actions. The combination of such properties attracts significant interest to the RL framework in the industry \cite{nian2020review} and other fields (e.g. risk management \cite{chakraborty19}, network traffic control \cite{arel2010reinforcement}, healthcare \cite{gottesman2018evaluating}) where the ability to look far ahead provides great benefits for decision making.

Thus, in this article, we report a successful implication of the RL algorithm to the production process on the metallurgical pickling line. We formulate our problem as a continuous optimal speed control task with the maximization of an average STU's speed over the long run as the main objective. 

\section{Process and problem description}

In Figure \ref{fig:schema} the scheme of the pickling line is presented. The movement of a steel strip within two units, the FTU and the TTU is characterized by clear periodicity. That periodicity is determined by the length of a particular strip. All steel strips processed on the pickling line come in a rolled state. As soon as the head of a new strip is welded with the tail of the previous one, the unrolling process starts. The FTU moves evenly and then slows down when the strip length left in the decoiler becomes sufficiently small (that is defined by lines settings). When the unrolling process is finished the welding process starts again, and during it the whole FTU is immovable. Then the process repeats for the next coil. The same is fair for the TTU, where instead of the welding process the cutting process takes place, and instead of residual length in the decoiler, a residual length before strip blades is used. The duration of the welding or cutting stages has some variability but can be predicted with acceptable errors.

The continual movement of STU becomes possible due to the presence of two loopers, one between the FTU and the STU, the other one between the STU and the TTU, and a separate operation of every unit. The amount of steel stored in every looper is controlled by the speed difference of the units. The excess speed of the FTU over the STU results in the fact that the amount of steel entering the first looper is higher than the amount of steel exiting it. Thus, the looper grows in volume to the resulting difference. During the welding stage, the excessive amount of steel stored in the looper can be utilized for uninterrupted movement. The opposite situation is for the STU and TTU. When the cutting stage takes place, the amount of steel processed during the stage is stored in the second looper. Then, the excessive amount of steel is reduced in the following stage by applying the speed of the TTU higher than the STU speed. The loopers are limited in capacity (have lower and upper bounds) and the capacity of the second looper is lower than the capacity of the first one. When the amount of steel stored in the first looper approaches the upper bound (or the lower bound for the second looper) the speed of the corresponding unit is forced to the speed of the STU. At the same time, the crossing of the opposite boundary can lead to long downtime.

Therefore, we formulate the entire problem as follows: 
\begin{itemize}
\item parameters: $L_f$ - strip length, unrolled in FTU; $L_t$ - strip length, recoiled in TTU; $\upsilon_f$ - speed of FTU; $\upsilon_t$ - speed of TTU, $t_W$ - welding time and $t_C$ - cutting time; 
\begin{equation}
  t_f = \frac{L_f}{\upsilon_f} + t_W
  \quad\text{and}\quad
  t_t = \frac{L_t}{\upsilon_t} + t_C
\end{equation}
\item decision variable: $\upsilon_s$ - speed of STU,
\begin{equation}
  \upsilon_s \leq \min (\upsilon_{s, set}^1, \dotsc, \upsilon_{s, set}^n)
\end{equation}
, where $\upsilon_{s, set}^i$ - the upper-speed boundary for every particular strip defined manually from the speed table, and $n$ - strips, currently in the STU of the pickling line;
\item constraints:
\begin{equation}
  V_f^{min} < \sum_{i=1}^{t_f} (\upsilon_{f, i} - \upsilon_{s, i})
\end{equation}
\begin{equation}
  \sum_{i=1}^{t_t} (\upsilon_{f, i} - \upsilon_{s, i}) < V_s^{max}
\end{equation}
, where $V_f^{min}$ - the boundary for the first looper; $V_s^{max}$ - the boundary for the second looper;
\item objective function:
\begin{equation}
  maximize  \sum_{i=1}^{n} \upsilon_{s, i}, {} n \to \infty
\end{equation}
\end{itemize}

\section{Solution approach}

\begin{figure}[ht]
  \centering
  \includegraphics[width=0.5\textwidth]{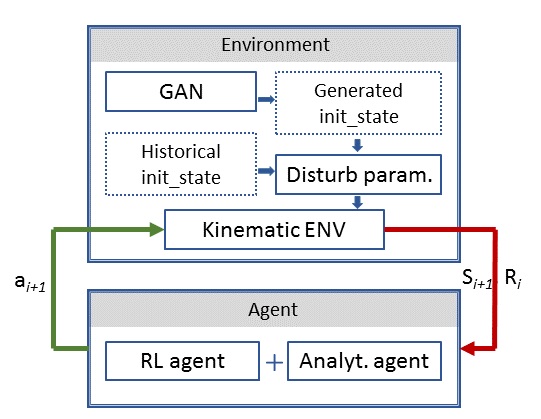}
  \caption{Scheme of solution}
  \label{fig:approach}
\end{figure}

\subsection{Scheme of solution}
The general scheme of the problem solution is shown in Figure \ref{fig:approach}. There are two principal parts in our approach: the environment reflecting the dynamics of the pickling line and the agent making the decision on the action to be taken. 

We developed a three-level environment with each level describing some particular elements of the system. The first level represents the steel strips that entered the line. The GANs (Generative Adversarial Network) have been used to simulate the real sequence of processed strips. The second level can be associated with the disturbance introduced by steel strips to the production mechanisms. That disturbance arises mainly from the variance in geometrical parameters of the strips and could be modeled by using the probabilistic approach. Finally, the third level (a digital twin or Kinematic ENV) characterizes the entire dynamic of the line by the principles described in the "Process and Problem Description" part. Methods of mathematical modeling have been exploited to simulate the behavior of separate units and to get a highly interpretable scheme of mutual influences.

We also used a two-step agent, consisting of a conservative agent and an RL (reinforcement learning) agent. By the conservative agent (\textsl{C-Agent}), we understand the analytical solution that calculates the speed every second during the production process. This agent uses mathematical modeling and the basics of mechanics to calculate the recommended speed. To build the RL agent, various deep RL algorithms were examined.

Hereinafter, we concentrate our attention on two parts of the approach, the development of the GAN algorithm producing the realistic tabular data characterizing the steel strip parameters and the investigation of the RL algorithm enabled to improve the solution suggested by the conservative agent. All the estimates in the article, except for those, in conclusion, were calculated using the Kinematic environment.  

\subsection{GAN: Input data generation}
The training and validation of a reinforcement learning system require a significant amount of input data that was not possible to acquire from real datasets. In our case, the input data represent steel strip characteristics of the metal passing through the pickling line. However, there was available a medium-sized historical dataset only, which could be either augmented or fed to a generative adversarial network (GAN).

Since GANs were introduced \cite{goodfellow2014}, their applications initially emerged in the tasks involving synthetic image generation and augmentation \cite{radford2015unsupervised, shangguan2019}. The homogeneous structure of images made them perfect objects for this pioneering approach, but recently an increasing number of researchers have been using GANs to create data of different structures, including time series, audio, video \cite{yoon2019time, donahue2018adversarial, vondrick2016generating}, and tabular data. For instance, GANs were successfully implemented to augment healthcare records \cite{che2017boosting}. Xu at el. used a Tabular GAN to generate tabular data with mixed variable types, using preliminary a Gaussian Mixture model for multi-modal columns and a long-short term memory network (LSTM) as a generator \cite{xu2018synthesizing}, or a Conditional Tabular GAN to model rows of data with discrete and continuous columns \cite{xu2019modeling}. Thus, to produce realistic synthetic data and to avoid limitations on the size of a synthesized dataset, we have chosen to use GANs.

For each epoch of the reinforcement learning system training, we required a short sequence of steel strips with certain parameters, thus the task was to produce batches of steel strip characteristics of length 15-20. The realistic sequence of steel strips is crucial in this task because timings of different technological processes vary from strip to strip depending on their parameters and on the combination of parameters of the adjacent pair of strips. Disturbance introduced by Kinematic ENV heavily depends on steel strip parameters. Table~\ref{tab:strips} shows the key parameters of the steel strip. 

\begin{table*}[ht]
  \caption{Basic Steel Strip Parameters}
  \label{tab:strips}
  \centering
  \begin{tabular}{p{0.3\textwidth}p{0.6\textwidth}}
    \hline
    Parameter&Characteristics\\
    \hline
    Steel grade & Categorical, over 100 unique values, highly imbalanced. Sequential properties are of great importance. \\
    Original width & Integer. Sequential properties are of great importance – the difference between consequential strips widths matters.\\
    Resulting width & Integer. Depending on the original width that might be adjusted on the pickling line. \\
    Thickness & Integer. Sequential properties are of great importance, there are limitations on thickness differences for adjacent strips.\\
    Weight & Integer, continuous. \\
    Coiling temperature & Float, continuous. \\
    Number of strips in resulting coil & Float. Strips may be rearranged to different coils on the pickling line. One strip can be cut into several strips and are coiled separately or 2-3 strips could be coiled together. \\
  \hline
\end{tabular}
\end{table*}

The steel grade is a category of steel comprising its chemical and mechanical properties and other parameters that are directly or indirectly conditioned on the steel grade, thus we have chosen to use a CGAN architecture to generate numerical parameters for this task and to use steel grade as a conditioning parameter. Therefore, we design a two-phase approach to model tabular data with sequential properties. 

\subsubsection{Steel grade generation with RNN}
Firstly, we generate a sequence of steel grades using an LSTM (Long short-term memory) recurrent neural network \cite{hochreiter1997long}. We have chosen this architecture because the parameter behaves as a sequence and can be generated using classical sequence generation approaches. Steel strips with the same parameters: grade, width, thickness - are usually organized in batches of different lengths during pickling and are pickled consequently. The size of the batch depends on the steel grade, for rare specialized grades the batches are usually smaller. Moreover, there might be consequential batches of the same steel grade, but with different other parameters. Thus, we add an $"END"$ token to the vocabulary of unique steel grades to denote the end of the batch. 
In this model called GradeGenerator, we used a single LSTM layer with 512 units, a dropout layer, a dense layer with softmax activation for the final output, and trained for 500 epochs optimizing the cross-entropy loss. A smaller number of hidden units turned out to be insufficient to capture specific patterns. We use the batch size of 256, sequences of length 20, and do not use an Embedding layer as the size of the unique steel grades vocabulary is relatively small. Then we sample novel sequences of steel grades according to the softmax distribution to make the results more diverse. 

\subsubsection{Conditional generative adversarial net (CGAN) for numeric parameters}
After that, we apply a conditional adversarial network algorithm using steel grade from GradeGenerator as an input. The Generative adversarial network consists of two neural networks: a generator $G(x)$ that creates the data and a discriminator $D(x)$ that tries to distinguish generated data from real data. Both are trained simultaneously through a minimax game. The discriminator tries to maximize the objective while the generator tries to minimize it, i.e to fool the discriminator by generating realistic samples. The generator uses the feedback provided by the discriminator to synthesize more and more believable samples. For the CGAN the loss is modified and the distributions of the variables are conditioned on $y$, in our case it is the steel grade $Gr$ resulting in the following objective function:
\begin{equation}
\begin{aligned}
\min_{G} \max_{D} V(D,G) = & \mathbb{E}_{x\sim p_{data}(x)}[log D(x|y)] + \\
                         & E_{z\sim p_{z}(z)}[log(1 - D(G(z)|y))] 
\end{aligned}
\end{equation}
The generator takes as an input the Gaussian noise vector of length $n$ with the probability density function $z = \frac{1}{{\sigma \sqrt {2\pi } }}e^{{{ - \left( {x - \mu } \right)^2 } \mathord{\left/ {\vphantom {{ - \left( {x - \mu } \right)^2 } {2\sigma ^2 }}} \right. \kern-\nulldelimiterspace} {2\sigma ^2 }}}$ with $\mu = 0, \sigma = 1$  along with the embedded steel grade. The dimension of the embedding layer $E(Gr)$ is equal to $n$. The architecture of the generator is the following:

\begin{itemize}
\item noise = Flatten(z ($\mu = 0, \sigma = 1$) )
\item label = Flatten(Embedding(Steel Grade))
\item product = Multiply(noise, label)
\item x1 = LeakyReLu(Dense(product, 256))
\item x2 = LeakyReLu(Dense(x1, 128))
\item x3 = LeakyReLu(Dense(x2, 64))
\item y = Reshape(\\
    LeakyReLu(Dense(x3, n*seqlength)))
\end{itemize}

\begin{figure}[ht]
  \centering
  \includegraphics[width=0.5\textwidth]{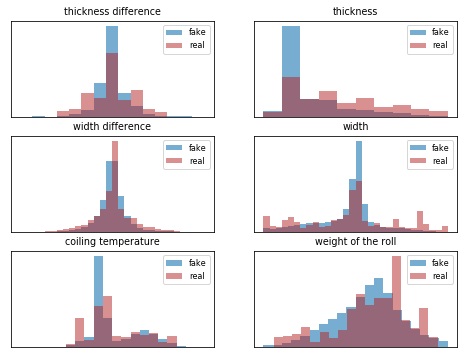}
  \caption{Comparison of real and generated strips}
  \label{fig:rolls}
\end{figure}

The discriminator shares the same architecture, with the difference that the last layer is a dense layer with one neuron. We treat the data as if it was a small image, which results in quite a good approximation of sequential properties. The real strip data are standardized before being fed to the discriminator. 
Then we train either the generator or the discriminator at a time with the weights of the other part frozen, using techniques proven to help achieve more stability in GAN training, for example, label smoothing \cite{salimans2016improved} and training the discriminator on $k$ times as much data as we give to the generator on each epoch \cite{goodfellow2016nips}. We use the Adam optimizer  \cite{kingma2014adam} and train a model for 2000 epochs. The results are presented in Figure \ref{fig:rolls}. The distribution of the crucial parameters of the coils is approximated quite well.  

\subsection{RL: Model description and parameters choice}
To solve our continuous optimization problem simply, we treat it as a sequence of episodic tasks, with the length of an episode limited by 20 consecutive steel strips passed through the kinematic environment described earlier. Therefore, one can transform the objective function to the following: $maximize  \sum_{i=1}^{n} \upsilon_{s, i}$, where  $n = \sum_{i=1}^{20} t_{f,i}$. Meaning the general approach widely accepted for the RL framework we define the following settings: Reward function, State, Policy evaluation strategy and Action.

We exploit an approach when two agents cooperatively define the finally applied action. The resulted action is obtained as a plain sum of actions evaluated by the agents and is determined as the speed of the STU at every time step.

\subsubsection{Reward}
Although we are using a two-agent approach, we consider the reward function to be applied only for the second agent (RL-agent). According to the problem description, there should be two main contributors to the reward function: a reward associated with an increase in the values of the objective function and penalties acquired from a violation of constraints. The analytical solution was built to keep the highest possible STU's speed in conditions that no constraints are broken. Therefore, any deviations from such safe behavior are considered to be the contribution of the RL agent and reflected in the reward function. We define the reward function of our RL agent consisting of the following terms: the contribution to the STU's speed with a positive sign, the violation of constraints associated with the death of the agent with a negative sign, and, optionally, the degree of proximity to boundaries of constrains. The term "death" for an agent hereinafter denotes the violation of constraints. 

\subsubsection{State}
The state vector was chosen as follows $$S_i = (V_{f,i}, V_{s,i}, \upsilon_{f,i}, \upsilon_{s,i}, \upsilon_{t,i}, t_{W, pred}, L_{f,i}, L_{t,i})$$, where $t_{W, pred}$ indicates the expected time left to the end of the welding stage (based on the predicted welding time value).

\subsubsection{Policy evaluation strategy}
We used Deep Q-learning for the evaluation of the most optimal STU's speed regulation strategy. In Q-learning the action-value (or Q-value) is used to estimate the expected sum of the gamma-discounted rewards acquired from taking action $a_i$ in state $s_i$ and yielding reward $r_{i+1}$ and a subsequent transition to state $s_{i+1}$. In terms of the \textit{Bellman equation} it could be rewritten for the immediate reward in the form \cite{sutton2018reinforcement}: 
\begin{equation}
\begin{aligned}
Q(s_i, a_i) \gets & Q(s_i, a_i) + \\
& \alpha [r_{i+1} + \gamma \max_{a} {Q(s_{i+1}, a)} - Q(s_i, a_i)]
\end{aligned}
\end{equation}

In Deep Q-learning instead of the scalar Q-value the Q-value function approximation is used with the neural network (NN) as a function approximation technique.

Considering the clear periodicity in the technological routine of the FTU and the TTU associated with the repetition of even motion, then uniformly accelerated motion and final immobility, one can redefine the behavior of these units in terms of stages. Therefore, one could distinguish three stages: boost ('0'), slowdown ('2') and welding (cutting) ('3'). Meaning the situation when the speed of the FTU or the TTU is forced to speed of the STU at the amount of steel in the looper approaching a certain boundary, one could add an extra synchronization stage ('1'). Thus, there are sixteen stage combinations. For each combination ($C^k, k=1...16$) we initialized a separate neural network responsible for the Q-value function approximation. Every network takes normalized state values as an input and passes them through the fully-connected network. Every hidden layer of the network is built from neurons with a ReLu activation function and an l1-regularizer. The last linear layer outputs a 1-d vector with Q-values associated with available actions, ranging from 1 to 10. While naturally speed is understood as a continuous quantity, in our case the simplification of it to the discrete variable is explained by the controller's regulation abilities. In addition to the stage combination parameter, we also define the active-time parameter ($T^k, k=1...16$): the k-th stage combination is active for the total time of $T^k$ during the length of one episode (with $\tau$=1...$T^k$).

\begin{algorithm}[htb]
\caption{Pseudo-code for Q-value function approximation learning}
\label{alg:pseudocode}
Initialization of kinematic ENV, \textsl{C-Agent}\;
Initialization of NN $Q(s, a| \theta)$ for every combination $C^k$\;
\For{episode in range(1, N)}{
    Receive coil sequence, disturbance parameters\;
    Receive initial state $s_1$\;
    \For{t in range(1, T)}{
    \If{$C_t^k \neq C_{t-1}^k$}{
    Observe state $s_{t,\tau}^k$ and reward $r_{\tau-1}^k$\;
    
    Set $z_{\tau-1} = r_{\tau-1}^k + \gamma \max_{a} {Q^k(s_{t,\tau}^k, a_{\tau-1}^k | \theta)}$\;
    Update weights of NN of this stage combination $C_t^k$ by minimizing the loss:
    $L(\theta) = \frac{1}{N} \sum_{\tau} {|z_{\tau-1} - Q^k(s_{\tau-1}^k, a_{\tau-1}^k | \theta)|^2}$\;
    Update step size $\alpha$ to reduce on factor 0.003;
    
    Select action $a_{\tau,RL-agent}^k$ according to $\epsilon$-greedy policy from $Q^k(s_{t,\tau}^k, a_{\tau-1}^k | \theta)$\;
    Select action $a_{\tau,C-agent}^k$\;
    Receive cumulative action $a_{\tau}^k = a_{\tau,RL-agent}^k + a_{\tau,C-agent}^k$\;
    
    }
    Execute action $a^k$ and observe state $s_{t+1}$\;
    Update reward $r^k$ until stage combination ends\;
    \If{constraints were violated}{
    receive reward
    \textbf{break}\;}
    }
}
\end{algorithm}

Switching between different NN could be described as follows when a new stage combination becomes active the appropriate NN is taken (at time $\tau$) and the Q-value is updated with the reward evaluated from previous steps (at the time ($\tau-1$)) with the same function approximation network. The update for the function approximation as well as a choice of the appropriate action is done only once at the beginning of a new stage combination and is kept until the next stage combination arrives.
The pseudo-code for the entire learning procedure is shown in Algorithm~\ref{alg:pseudocode}

\begin{table*}[!ht]
  \caption{Setting's variations for RL-agents}
  \centering 
  \label{tab:actions}
  \begin{tabular}{p{0.3\textwidth}p{0.3\textwidth}p{0.3\textwidth}}
    \hline
    Parameter & \textsl{P-Coop Agent} & \textsl{F-Coop Agent}\\
    \hline
    The number of NNs layers & 2 (8 | 8) & 3 (32 | 64 | 16) \\
    The state vector & $(V_{f,i}, V_{s,i}, \upsilon_{f,i}, \upsilon_{t,i}, t_{W,pred},$ $ L_{f,i}, L_{t,i})$ & $(V_{f,i}, V_{s,i}, t_{W,pred}, L_{f,i}, L_{t,i})$ \\
    The reward contributors & (action, constraints violation) & (action, constraints violation, proximity to boundaries of constraints) \\
    The reward on stage calculation ($t'$ - stage combination duration) & $\sum_{i=1}^{t'} r_i$ & $\frac{1}{t'} \sum_{i=1}^{t'} r_i$\\
    Action from agent & [0, 9] & [-5, 5] \\
    Frequency & Every stage combination, except where welding or cutting stage is involved & Every stage combination, plus every 30 sec. during combination time \\
  \hline
  \end{tabular}
\end{table*}

\subsubsection{Action, cooperation strategy}
Next, we introduce two RL agents with several different key points in respect to their interaction with the conservative agent, while the architecture of the solution stays the same.

We explore the behavior of two RL agents since they represent a different approach to the final solution. On the one hand, we consider a fully cooperative agent (\textsl{F-Coop Agent}), when the decision on the action applied to the system is assessed by both agents at every change of the stage combination. On the other hand, we describe a more conservative but more industry-tolerated agent (\textsl{P-Coop Agent}), where the cooperative decision is taken for all stage combinations except those where the welding or cutting stage is involved. That is done because the duration of these stages has high variance and any lapses in logic may result in continuous downtime of the pickling line. 

The variations in the earlier introduced solution done for every agent are shown in Table~\ref{tab:actions}. The modified parameters are the number of layers in NNs responsive for the Q-value approximation, the composition of the state vector, the reward function evaluation strategy, the range of available actions, and the frequency of actions selection.

\section{Results and Discussion}

The agents were firstly trained at 800 episodes with generated initial conditions (IC) and $\epsilon$-greedy policy. Then $\epsilon$ was forced to zero and the training was continued to the next 200 episodes with the IC taken from the historical data. All the IC and states were precomputed to ensure the re-productivity and identity of the starting condition for different agents. For generated IC the pipeline shown in Figure \ref{fig:approach} was used. Below, the results obtained at the last 100 episodes with historical IC are discussed.

\begin{figure*}[ht]
  \centering
  \includegraphics[width=\textwidth]{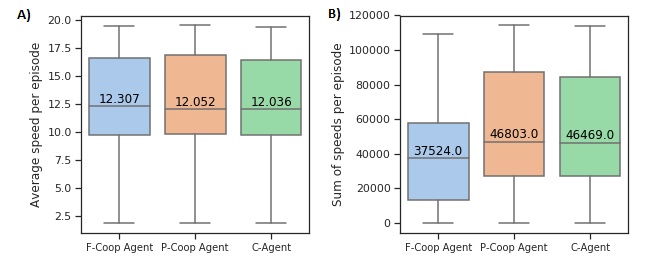}
  \caption{Objective function for \textsl{F-Coop Agent}, \textsl{P-Coop Agent}, and \textsl{C-Agent} agents. A) Average STU speed per episode; B) Sum of STU speed during the episode. Values shown in boxes correspond to median values of distributions.}
  \label{fig:objective}
\end{figure*}

Above we have determined the objective function of our problem as a sum of the STU speeds obtained at every time step during the episode. With the limitation of the episode to the production time of 20 consecutive steel stripes one could redefine the objective function as an average STU speed per episode. In Figure \ref{fig:objective} the results for the original objective function as well as for the modified one (average speed) are shown. Here and further we distinguish the \textsl{C-Agent} applied stand-alone as a baseline solution to compare the gain in productivity raised from the implication of the RL-approach and the \textsl{C-Agent per stage} as an agent that contributes to the decision made by the RL-agent. The \textsl{C-Agent}, as a baseline solution, acts every second, while the RL agents are more restricted in the acting frequency (see Table~\ref{tab:actions}). 

As it is seen from Figure \ref{fig:objective} the decision on which the agent shows the best strategy depends on the approach taken for the objective function. The \textsl{F-Coop Agent} shows the best performance if an average speed is considered as a decision function and significantly loses when the sum of speed per episode is assessed. Such behavior could be explained by the rate of agents death. The early death of the agent results in a shortening of the episode length, while its effect on the average speed is not obvious and even may increase the average speed value. 

\begin{table*}[ht]
  \caption{Amount of deaths for different agents}
  \centering
  \label{tab:death}
  \begin{tabular}{l p{0.018\textwidth} p{0.018\textwidth} p{0.018\textwidth} p{0.018\textwidth} p{0.02\textwidth} p{0.018\textwidth} p{0.018\textwidth} p{0.018\textwidth} p{0.018\textwidth}}
    \hline
   Agents & \multicolumn{9}{c}{Stage combinations}\\
    & '01' & '03' & '10' & '13' & '23' & '30' & '31' & '32' & '33'\\
    \hline
    \textsl{F-Coop Agent} & 5 & 3 & 1 & 6 & 0 & 3 & 2 & 2 & 22 \\
    \textsl{P-Coop Agent} & 0 & 1 & 1 & 2 & 1 & 2 & 1 & 2 & 15 \\
    \textsl{C-Agent} & 0 & 1 & 1 & 2 & 1 & 2 & 2 & 2 & 15 \\
  \hline
  \multicolumn{10}{c}{Stage combinations: 'stage in FTU' + 'stage in TTU'}
\end{tabular}
\end{table*}

\begin{figure*}[htb]
  \centering
  \includegraphics[width=\linewidth]{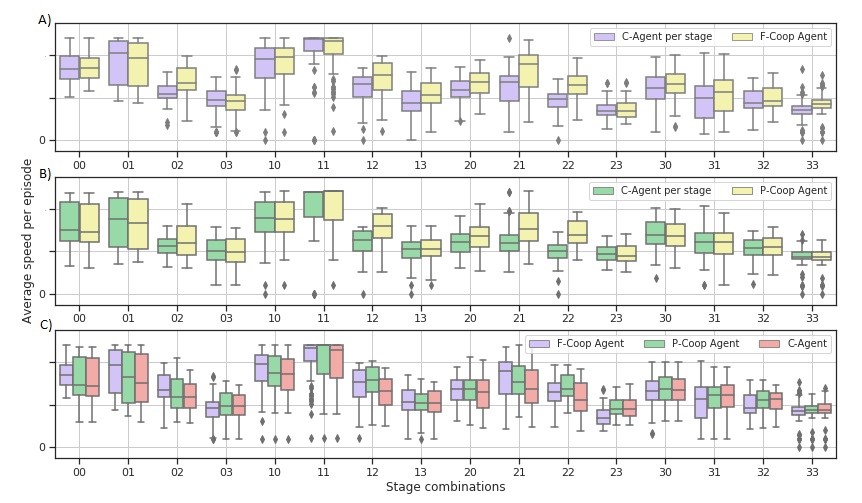}
  \caption{The average STU speed over different stage combinations. A) \textsl{F-Coop Agent} vs. \textsl{C-Agent per stage}; B) \textsl{P-Coop Agent} vs. \textsl{C-Agent per stage}; C) \textsl{F-Coop Agent} and \textsl{P-Coop Agent} vs. \textsl{C-Agent}.}
  \label{fig:speed_by_stage}
\end{figure*}

In Table~\ref{tab:death} the amount of agents deaths is presented with the differentiation for the stage combination when the death happened. Logically, most of the deaths occurred at the stage combinations when either welding or cutting stage('3') took place. These stages possess a high variability in the duration time and while this time could be predicted with some errors the agents hardly adapted to such variations. Moreover, some deaths could be predetermined by starting conditions, when the initiation took place at a stage combination with welding or cutting stage involved, death could be expected with a high probability. Nevertheless, the comparison of the excessive death rate of the RL-agents (Fully Cooperative and Partly Cooperative) over the \textsl{C-Agent} (27\% death rate) shows a significant growth of deaths rate for the \textsl{F-Coop Agent} (43\% death rate) over the \textsl{P-Coop Agent} (26\% death rate). These data also confirm the assumption done from the objective function consideration, that lower values of the speed's sum for the \textsl{F-Coop Agent} as well as for the \textsl{C-Agent} could be attributed to a higher death rate.

\begin{figure*}[ht]
  \centering
  \includegraphics[width=\linewidth]{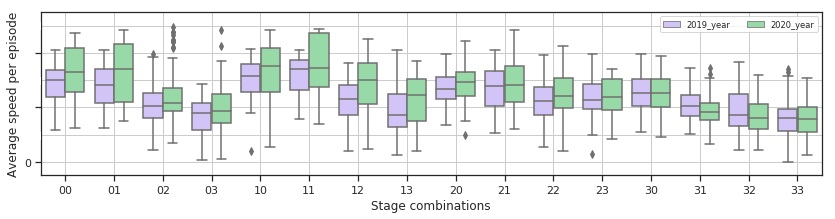}
  \caption{Average STU speed over different stage combinations, the factual data collected before (2019 year) and after (2020 year) the implementation of the suggested approach. Data were collected according to the logic described in the manuscript, 100 runs of 20 consecutive steel stripes are used as input data for the plot.}
  \label{fig:history}
\end{figure*}

To clarify the reasons for the excessive death rate the comparison of the speed taken by the agents with the speed contributed by the \textsl{C-Agent per stage} is done for every stage combination (Figure \ref{fig:speed_by_stage}). Presented data is obtained as follows, the average STU speed over different stage combinations per episode is calculated, then data from all 100 episodes is aggregated; therefore, one box at the plot represents the distribution of the average STU speed for a given stage combination at 100 separate runs. Since we know that most of the death is associated with having-'3' combinations we could take a more close look at them. According to the initial descriptions of the agents, the \textsl{P-Coop Agent} is restricted at having-'3' stages by the decision made by the \textsl{C-Agent per stage}. Therefore, there is only a minor deviation, less than 3.5\%, of the speed displayed by the agent over the basic decision of the \textsl{C-Agent per stage} (Figure \ref{fig:speed_by_stage} (B)). Meanwhile, the \textsl{F-Coop Agent} could act with positive or negative additive over the base decision of the \textsl{C-Agent per stage}. As it is shown in Figure \ref{fig:speed_by_stage} (A) the final action taken by the \textsl{F-Coop Agent} in having-'3' stages exceeds the speed suggested by the \textsl{C-Agent per stage} in most of the cases. Thus, from 7 having-'3' stages in 5 of them, we detect a surplus, and in 3 of them, the surplus is over 15\% of the basic decision of the \textsl{C-Agent per stage}. Such speed advantage of the \textsl{F-Coop Agent} over the \textsl{C-Agent per stage} despite of the high death rate of the first one could be explained by two reasons. The first reason is the more risky behavior of the \textsl{F-Coop Agent} agent compared to the \textsl{C-Agent per stage}. There is an emergency system that forces the speed of the STU to decrease when the system is approaching a constraints violation. That emergency braking system was originally built for the \textsl{C-Agent} to hedge against uncertainty arising from errors in predicting the welding/cutting times. However, the \textsl{F-Coop Agent} exploits it to set high-speed values at the beginning of stage combination and at the end of the stage combination to go at the lowest possible speed, forced by the emergency braking system. We find confirmation of that behavior by analyzing the actions of the agent. However, the aforementioned errors in predicting the welding/cutting time also occur for the \textsl{F-Coop Agent}, apparently causing additional deaths. The second reason could be that the values of the reward function are chosen insufficiently well: if the penalties values are too small compared to the reward values, then the agent will set high-speed values for the stage combination. However, in our case, the penalty was more than two orders of magnitude higher than the reward, so we believe that the first reason (i.e. risky behavior) is more likely to explain the speed advantage of the \textsl{F-Coop Agent}. 

\begin{table*}[ht]
  \caption{Speed excess of RL-agents in percent over the conservative agents}
  \centering
  \label{tab:surplus}
  \begin{tabular}{l p{0.01\textwidth} p{0.01\textwidth} p{0.01\textwidth} p{0.01\textwidth} p{0.01\textwidth}}
    \hline
   Agents & \multicolumn{5}{c}{Stage combinations}\\
    & '02' & '12' & '20' & '21' & '22'\\
    \hline
    \textsl{F-Coop Agent} vs. \textsl{C-Agent per stage} & 23 & 17 & 16 & 30 & 36\\
    \textsl{F-Coop Agent} vs. \textsl{C-Agent} & 14 & 17 & 6 & 31 & 18\\
    \textsl{P-Coop Agent} vs. \textsl{C-Agent per stage} & 7 & 26 & 12 & 27 & 35\\
    \textsl{P-Coop Agent} vs. \textsl{C-Agent} & 1 & 21 & 6 & 12 & 25\\
  \hline
\end{tabular}
\end{table*}

Besides the having-'3' stages, the excess in the speed of the agents over the speed suggested by the \textsl{C-Agent per stage} occurred also for having-'2' stages. In Table~\ref{tab:surplus} the surplus as the percentage of the base agent speed is calculated. Both agents, the \textsl{F-Coop Agent}, and the \textsl{P-Coop Agent} show identical behavior that could be attributed to the potentially positive effect of the RL approach. In Figure \ref{fig:speed_by_stage} (C) the final results of the comparison of the speeds from the \textsl{F-Coop Agent}, the \textsl{P-Coop Agent}, and the \textsl{C-Agent} are presented. The same tendency for having-'2' stages is preserved here, we detect a distinct excess of the RL-agents speed over the conservative \textsl{C-Agent}, although it is less than in the case of the \textsl{C-Agent per stage} (Table~\ref{tab:surplus}). Therefore, we could consider it as the main effect of the application of the RL agents to the current problem. 

From all of the above we could conclude that while the \textsl{F-Coop Agent} shows sometimes better STU speed values than the \textsl{P-Coop Agent} and could be considered as a more flexible agent, its risky decisions are responsible for a higher death rate and therefore for a lower value of the objective function. At the same time, the more conservative \textsl{P-Coop Agent} allows the increase in speed of the STU only in a safe condition and thereby results in higher productivity. 

\section{Conclusions}

In this work, we have proposed a multi-level approach addressed to the speed line control problem on a metallurgical pickling line. The main objective of STU's speed regulation is to maximize the productivity of the pickling line, on the one hand, and, on the other hand, to provide the conditions when all quality requirements for the processed stripes are met. At the first level of our approach, the combination of LSTM and CGAN has been exploited to generate the realistic input data and thereby overwhelm the deficiency in original data necessary to solve the problem. Then several steps have been done, that are only partially described in the manuscript, such as the development of a C-agent based on the analytical solution of the problem and the kinematic environment simulating the real processes on the pickling line. Finally, at the third level, the complete architecture of the Reinforcement Learning system has been built and trained. The entire approach with the \textsl{P-Coop Agent} has been successfully applied at the Cherepovets Steel Mill and has allowed to increase the productivity of the pickling line by more than 6.5\% and to significantly improve line automation processes. In Figure \ref{fig:history} the comparison of the average STU speed over different stage combinations before and after the implementation of the entire approach is presented. Moreover, an additional quality check of the processed strips was done as well, which did not show a statistically significant difference in the number of over- and under-pickling defects before and after the solution deployment.The whole approach was released in two steps: firstly a base C-Agent (5\% productivity increase) and then the RL-Agent (1.5\% productivity increase). These numbers correspond to the factual increase in productivity of the pickling line in 2020. We used a technological stack including Kubernetes, Docker, Kafka, and OPC UA. All modelling was performed in Python. Later, the approach has been partially extended to one more unit at the Cherepovets Steel Mill. The Kinematic ENV has also been successfully used to test hypotheses regarding the productivity of the pickling line and has the potential to be used for scheduling problems. 

\section*{Acknowledgment}
The authors wish to thank the Flat-Rolled Products Subdivision at the Cherepovets Steel Mill and especially Andrei Fedotov and Alexandr Ruban for valuable knowledge and assistance during the project.

\bibliographystyle{elsarticle-num-names} 
\bibliography{references}





% \end{thebibliography}
\end{document}